\documentclass[10pt]{JournalTitle}
\usepackage{graphicx}
\usepackage[singlelinecheck=off,font={footnotesize,bf},labelsep=space]{caption}
\usepackage{cite}
\usepackage{algorithmic}
\usepackage{graphicx}
\usepackage{textcomp}
\usepackage{xcolor}
\usepackage{booktabs}
\usepackage{float}
\usepackage{amsmath}
\usepackage{amssymb}
\usepackage[cspex,bbgreekl]{mathbbol}
\usepackage[T1]{fontenc}
\usepackage[utf8]{inputenc}
\usepackage{lmodern}  
\usepackage{multirow}

\def\BibTeX{{\rm B\kern-.05em{\sc i\kern-.025em b}\kern-.08em
    T\kern-.1667em\lower.7ex\hbox{E}\kern-.125emX}}

\begin{document}
\title{TACFN: Transformer-based Adaptive Cross-modal Fusion Network for Multimodal Emotion Recognition}

\titleshort{Short Title on the Header}


\author{
	Feng Liu\footnotemark[1],\footnotemark[4],\footnotemark[5],
	Ziwang Fu\footnotemark[2],
        Yunlong Wang\footnotemark[3],
        Qijian Zheng\footnotemark[1]
        }


\footnotetext[1]{School of Computer Science and Technology, East China Normal University, Beijing 100084, China}
\footnotetext[2]{MTlab, Meitu (China) Limited, Beijing 100876, China}
\footnotetext[3]{Institute of Acoustics, University of Chinese Academy of Sciences, Beijing 100084, China}
\footnotetext[3]{Institute of Acoustics, University of Chinese Academy of Sciences, Beijing 100084, China}
\footnotetext[5]{Address correspondence to Feng Liu, lsttoy@163.com}


\doi{https://doi.org/10.26599/AIR.2023.9150019}

\volume{2}
\date{July 2023}
\page{9150019}

\abstract{The fusion technique is the key to the multimodal emotion recognition task. Recently, cross-modal attention-based fusion methods have demonstrated high performance and strong robustness. However, cross-modal attention suffers from redundant features and does not capture complementary features well. We find that it is not necessary to use the entire information of one modality to reinforce the other during cross-modal interaction, and the features that can reinforce a modality may contain only a part of it. To this end, we design an innovative Transformer-based Adaptive Cross-modal Fusion Network (TACFN). Specifically, for the redundant features, we make one modality perform intra-modal feature selection through a self-attention mechanism, so that the selected features can adaptively and efficiently interact with another modality. To better capture the complementary information between the modalities, we obtain the fused weight vector by splicing and use the weight vector to achieve feature reinforcement of the modalities. We apply TCAFN to the RAVDESS and IEMOCAP datasets. For fair comparison, we use the same unimodal representations to validate the effectiveness of the proposed fusion method. The experimental results show that TACFN brings a significant performance improvement compared to other methods and reaches the state-of-the-art. All code and models could be accessed from https://github.com/shuzihuaiyu/TACFN.}

	\maketitle
	\begin{multicols}{2}

		\firstletter{I}
n recent years, the proliferation of multimedia data, including short videos and movies, has propelled the rise of multimodal emotion recognition as a burgeoning field of study, eliciting significant interest among researchers and practitioners alike \cite{zhao2021emotion,poria2020beneath}. The objective of this task is to classify human emotions from a video clip using three modalities: visual, audio, and text. Multimodality offers a wealth of information that surpasses unimodality and better aligns with human performance behavior \cite{gan2017multimodal, nguyen2018deep}. Humans tend to perceive the world by processing and fusing high-dimensional inputs from multiple modalities simultaneously \cite{smith2005development}. However, how to efficiently extract and integrate information from all input modalities so that the machine can correctly recognize emotions remains a challenge for this task. On the one hand, the modality input representations are different, and efficient intra-modal encoding integrates high-level semantic features of the current modality. On the other hand, the learning between modalities is dynamically changing \cite{wang2020makes}, where some modal streams will contain more task information than others. The goal of this paper is to propose an innovative multimodal fusion approach.

In the existing research, multimodal fusion approaches can be classified as: early fusion \cite{10.1145/2070481.2070509,inproceedings2013early}, late fusion \cite{article2016late, 80193012017late} and model fusion \cite{sahay2020low, rahman2020integrating, yu2021learning}. Early fusion strategies involve fusing the shallow inter-modal features and focusing on mixed-modal feature processing while late fusion strategies involve finding the confidence level of each modality and then coordinating them to make joint decisions. With the development of deep learning, model fusion has significantly improved the performance of multimodal tasks compared to the previous two approaches. Model fusion enables flexible choice of fusion locations, taking into account the intrinsic correlation between sequence elements from different modalities \cite{hazarika2020misa}. Recently, due to the popularity of Transformer \cite{10.5555/3295222.3295349} in multimodal machine learning tasks, model fusion is often done using a transformer-based approach for different modal interactions. Typically, both SPT \cite{cheng2021multimodal} and PMR \cite{lv2021progressive} use cross-modal attention to perform fusion of multimodal sequences. Their approaches achieve reinforcement of the target modality by learning directed pairwise attention between cross-modal elements. Moreover, according to recent findings \cite{nagrani2021attention}, the cross-modal attention performance in fusion of high-level semantics is still able to excel in different tasks.

\begin{figure*}[!t]
\centering
\vspace{-0.5em}
\includegraphics[scale =0.6]{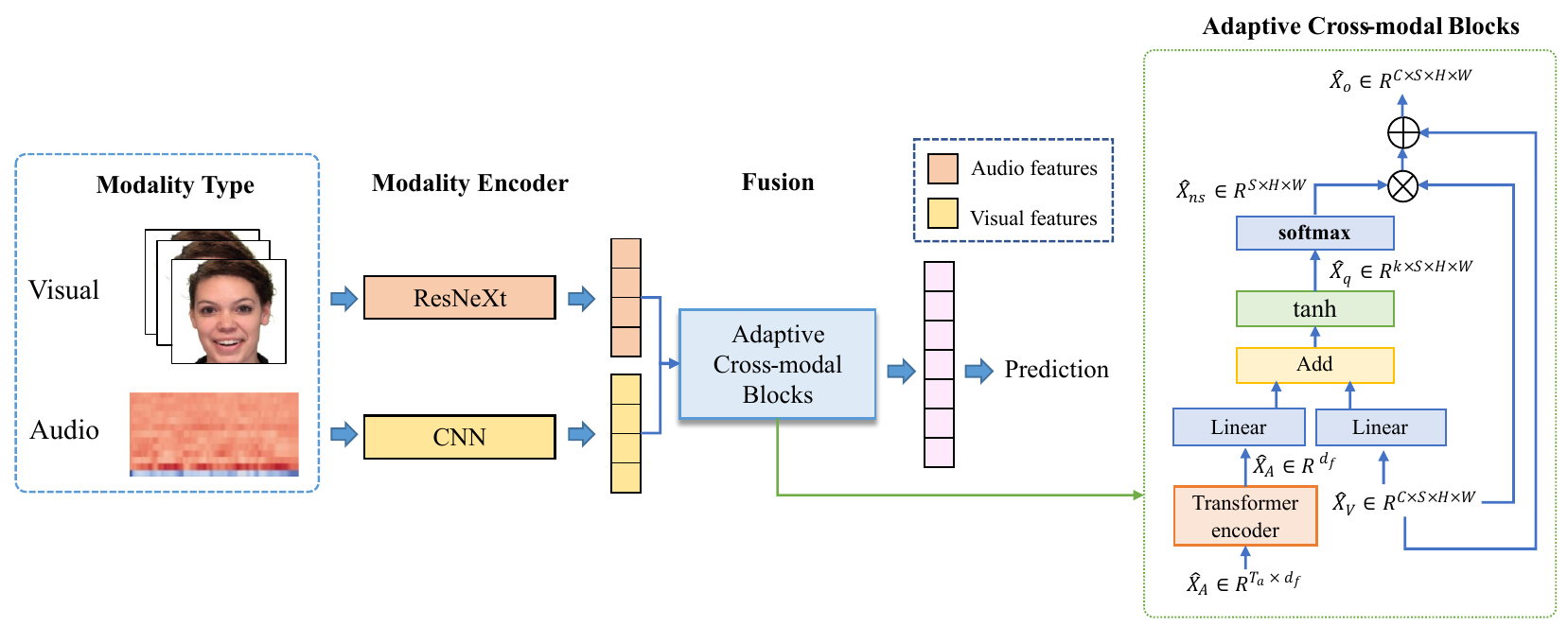}
\caption{The overall architecture of TACFN. \textbf{Left}: the flow structure of the whole model, divided into two parts: unimodal representation and multimodal fusion. \textbf{Right}: the adaptive cross-modal fusion block.}
\vspace{-0.5em}
\label{fig:framework}
\end{figure*}

Cross-modal attention, however, is challenged by the presence of feature redundancy.  Our analysis reveals that the audio and visual inputs are densely packed with fine-grained information, much of which is redundant.  During the fusion of audio and visual modalities, the paired cross-modal attention is quadratically complex in relation to the length of the multimodal sequence, making this operation inefficient.  Moreover, the cross-modal attention does not capture complementary features well.  For example, the audio modality is not entirely helpful for the visual modality, and the audio features that often reinforce the visual features may contain only a fraction of them.   Additionally, the information density of modalities for the emotion recognition task varies, with some modal flows carrying more task information than others.  For instance, visual modalities perform better in classifying ``happy" emotions, while audio modalities are more effective in classifying ``fearful" emotions.

To this end, we propose an innovative transformer-based adaptive cross-modal fusion network (TACFN) for multimodal emotion recognition. Specifically, we divide the network into two steps: unimodal representation and multimodal fusion. Firstly, for unimodal representation, we perform representation learning on the spatio-temporal structure features \cite{xie2017aggregated} of video frame sequences and MFCC features \cite{neverova2015moddrop} of audio sequences, respectively. Then, for multimodal fusion, the two core issues of reducing redundant features and enhancing complementary features are mainly considered. For reducing redundant features, a self-attention mechanism enables a modality to perform intra-modal feature selection, and the selected features will be able to interact with another modality adaptively and efficiently. To enhance complementarity, we fuse the selected modality with another modality by splicing and generating a weight vector.  The weight vector is then multiplied with another modality to achieve feature enhancement. Finally, we obtain the prediction of emotional categories by splicing the fused representation. For simplicity, this paper first focuses on the fusion of two modalities, visual and audio. Further, we extend the designed fusion block to cross-modal fusion of three modalities, visual, audio and text. We apply TACFN to the RAVDESS \cite{livingstone2018ryerson} and IEMOCAP \cite{iemocap} datasets, and the experimental results show that our proposed fusion method is more effective. Compared with other methods, our method brings a significant performance improvement in emotion recognition by the fusion strategy based on the same unimodal representation learning. Our method achieves the state-of-the-art.

The main contributions of this paper can be summarized as follows:

\begin{itemize}
    \item We propose an innovative transformer-based adaptive cross-modal fusion network (TACFN) for multimodal emotion recognition.
    \item We divide TACFN into two parts: unimodal representation and multimodal fusion. We perform intra-modal feature selection through a self-attention mechanism and achieve modal feature reinforcement by introducing a fused weight vector. This operation reduces redundant features and enhances inter-modal complementary information.
    \item We apply TACFN to the RAVDESS and IEMOCAP datasets. The experimental results show that the adaptive cross-modal fusion block brings a significant performance improvement to the state-of-the-art based on the same unimodal representation learning.
\end{itemize}

\section{Related Work}
Multimodal emotion recognition understands various human emotions by collecting and processing information from different modalities \cite{nguyen2018deep, nguyen2017deep}. This task requires fusion of cross-modal information from different temporal sequence signals. We can classify multimodal fusion according to the way it is performed: early fusion \cite{10.1145/2070481.2070509,inproceedings2013early}, late fusion \cite{article2016late, 80193012017late} and model fusion \cite{sahay2020low, rahman2020integrating, yu2021learning}. Previous work has focused on early fusion and late fusion strategies. Early fusion \cite{10.1145/2070481.2070509,inproceedings2013early} is to extract and construct multiple modal data into corresponding modal features before stitching them together into a feature set that integrates individual modal features. On the contrary, late fusion \cite{10.1145/2070481.2070509,inproceedings2013early} is to find out the credibility of individual models and then to make coordinated, joint decisions. Although these methods obtain better performance than learning from a single modality, they do not explicitly consider the intrinsic correlations between sequence elements from different modalities, which are crucial for effective multimodal fusion. For multimodal fusion, a good fusion scheme should extract and integrate valid information from multimodal sequences while maintaining the mutual independence between modalities \cite{hazarika2020misa}.

With the popularity of Transformer \cite{10.5555/3295222.3295349}, model fusion is gradually using Transformer to achieve multimodal fusion. Model fusion \cite{sahay2020low, rahman2020integrating, yu2021learning} allows flexible choice of fusion locations compared to the previous two approaches, taking into account the intrinsic correlation between sequence elements from different modalities. Typically, MulT \cite{tsai2019multimodal}, PMR \cite{lv2021progressive} and SPT \cite{cheng2021multimodal} all use cross-modal attention to model multimodal sequences. The cross-modal attention operation uses information from the source modality to reinforce the target modality by learning directed pairwise attention between the source modality and the target modality. Moreover, MMTM \cite{joze2020mmtm} module allows slow fusion of information between modalities by adding to different feature layers, which allows the fusion of features in convolutional layers of different spatial dimensions. MSAF \cite{su2020msaf} module splits each channel into equal blocks of features in the channel direction and creates a joint representation that is used to generate soft notes for each channel across the feature blocks. MBT \cite{nagrani2021attention} restricts the flow of cross-modal information between latent units through tight fusion bottlenecks, that force the model to collect and condense the most relevant inputs in each modality.

This work proposes a new attention-based adaptive multimodal fusion network that performs adaptive intra-modal selection through a self-attention mechanism to reduce its own redundant features and improve the ability to capture complementary features.

		\section{Methodology}
  In this paper, we divide TACFN into two steps, unimodal representation and multimodal fusion. Our goal is to perform efficient cross-modal fusion from multimodal sequences, aggregating intra- and inter-modal features to achieve correct emotion classification. Figure \ref{fig:framework} shows the overall framework.
\subsection{Modality encoding}

\subsubsection{Audio Encoder}

For the audio modality, recent work \cite{neverova2015moddrop, wang2020speech} has demonstrated the effectiveness of deep learning methods based on Mel Frequency Cepstrum Coefficient (MFCC) features. We design a simple and efficient 1D CNN to perform MFCC feature extraction. Specifically,  we use the feature preprocessed audio modal features as input, denoted as $X_{a}$. We first pass the features through a 2-layer convolution operation to extract the local features of adjacent audio elements. After that, we use the max-pooling to downsample, compress the features, and remove redundant information. The specific equation is as follows:
\begin{equation}
\begin{split}
    \hat{X}_{a} = {\rm BN}({\rm ReLU}({\rm Conv1D}(X_{a}, k_{a}))) \\
    \hat{X}_{a} = {\rm ReLU}({\rm Conv1D}(\hat{X}_{a}, k_{a}))
\end{split}
\end{equation}
\begin{equation}
    \hat{X}_{a} = {\rm Dropout}({\rm BN}({\rm MaxPool} (\hat{X}_{a})))
\end{equation}
where $BN$ stands for Batch Normalization, $k_{a}$ is the size of the convolution kernel of modality audio and $\hat{X}_{a}$ denotes the learned the high-level semantic features. Finally, we flatten the obtained features:
\begin{equation}
    \hat{X}_{a} = {\rm Flatten}({\rm BN}({\rm ReLU}({\rm Conv1D}(\hat{X}_{a}, k_{a}))))
\end{equation}

\subsubsection{Visual Encoder}

Video data are dependent in both spatial and temporal dimensions, thus a network with 3D convolutional kernels is needed to learn facial expressions and actions. We consider both the performance and training efficiency of the network and choose the 3D ResNeXt \cite{xie2017aggregated} network to obtain the spatio-temporal structural features of visual modalities. ResNeXt proposes a group convolution strategy between the deep segmentation convolution of ordinary convolutional kernels, and achieves a balance between the two strategies by controlling the number of groups with a simple structure but powerful performance. We use feature preprocessed visual modal features as input, denoted as $X_{v}$. We obtain the high-level semantic features of visual modalities by this network:

\begin{equation}
    \hat{X}_{v} = {\rm ResNeXt 50}(X_{v}) \in \mathbb{R}^{C \times S \times H \times W}
\end{equation}
where $\hat{X}_{v}$ denotes the learned semantic features, $C$, $S$, $H$ and $W$ are the number of channels, sequence length, height, and width, respectively.

\subsection{Fusion via cross-modal attention}

The cross-modal attention operation uses information from the source modality to reinforce the target modality by learning directed pairwise attention between the source modality and the target modality \cite{tsai2019multimodal, lv2021progressive}. Cross-modal attention is a modification of self-attention, with Q as one modality and K and V as another modality to obtain reinforcement of the modality. We define the cross-modal attention of two tensors $X$ and $Y$, where $X$ forms the query and $Y$ forms the keys and values used to reweight the query as ${\rm MCA}(X, Y) = {\rm Attention}(W^Q X, W^K Y, W^V Y)$. 

We take the obtained unimodal high-level representations and perform cross-modal attention interactions between audio and visual modalities to obtain reinforcing features of each other. We first encode the input $\hat{X}_{m}$ of the multimodal sequence into the length representations as $h_m$, $m \in \{v,a\}$. After that, we feed $h_m$ into the cross-modal Transformer encoder. An encoder consists of a sequence of $L$ cross-modal transformer layers, and each cross-modal transformer layer consists of Multi-head Cross-modal Attention (MCA), Layer Normalization (LN) and Multilayer Perceptron (MLP) blocks applying residual connections. To this end, we define a Cross-Transformer layer as:

\begin{figurehere}
\centering
\includegraphics[width=0.95\linewidth]{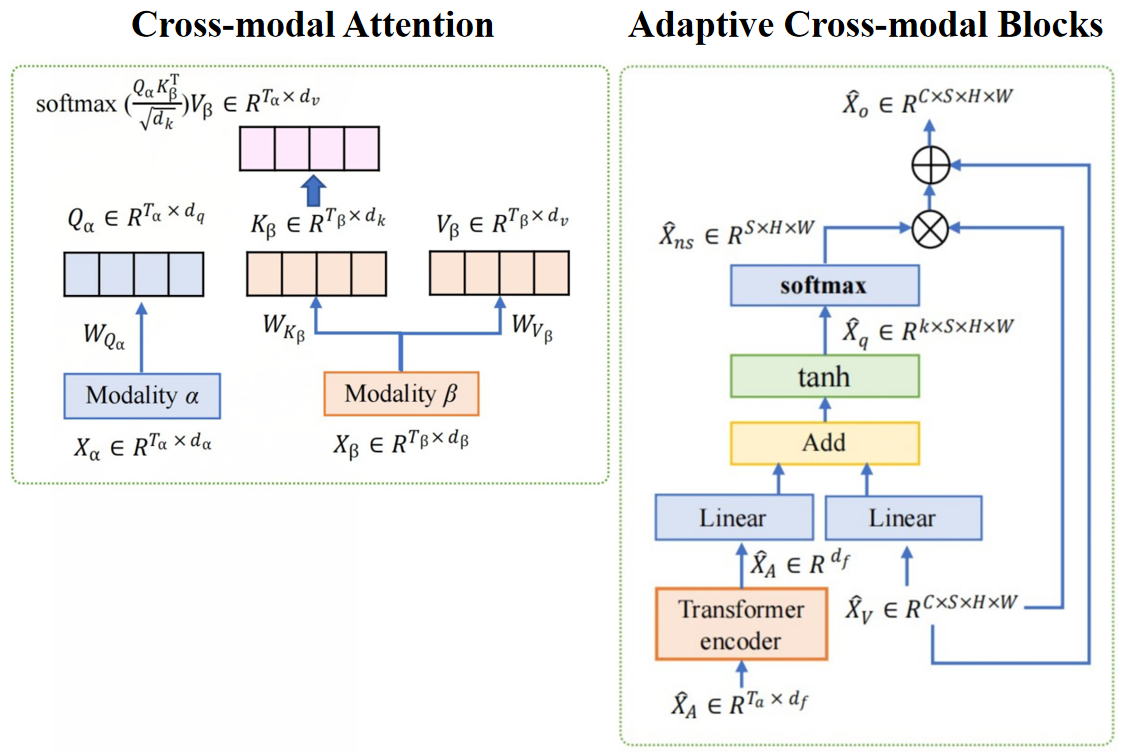}
\caption{Architectural elements of a cross-modal attention and an adaptive cross-modal blocks.}
\vskip 10.1pt
\label{fig:adaptive}
\end{figurehere}

\begin{equation}
    y^l = {\rm MCA}({\rm LN}(h_{\{v,a\}}^{l}), {\rm LN}(h_{\{a,v\}}^{l})) + h_{\{v,a\}}^{l},
\end{equation}
\begin{equation}
    h_m^{l+1} = {\rm MLP}({\rm LN}(y^{l})) + y^{l} \quad m \in \{v,a\}.
\end{equation}

The target modality is reinforced by encouraging the model to attend to crossmodal interaction between elements. The formula is as follows:

\begin{equation}
    h_a^{l+1} = {\rm Cross\raisebox{0mm}{-}Transformer}(h_a^{l}, h_v^{l}; \theta_{a})
\end{equation}
\begin{equation}
    h_v^{l+1} = {\rm Cross\raisebox{0mm}{-}Transformer}( h_v^{l}, h_{a}^{l}; \theta_{v})
\end{equation}
where $\hat{h}_a$ denotes the reinforced audio modality, $\hat{h}_v$ denotes the reinforced visual modality. Finally, we splice the reinforced modalities to obtain the fused data $I = [\hat{h}_a, \hat{h}_v]$.

\subsection{Fusion via adaptive cross-modal blocks}

The existing experiments show that model fusion further considers the internal relationship between modalities and has better effects and performance. However, there are some problems in the cross-modal fusion scheme adopted by the current model fusion, namely cross-attention:

(1) Feature redundancy exists in cross-modal attention.

(2) Cross-modal attention does not capture complementary features well.

(3) Since the modes change dynamically, some of them have more representation information for the task than others. For example, the visual modality classifies ``happy" better than the audio modality, and the audio modality classifies ``fearful" better.

Therefore, on the basis of Cross-modal, we introduce the design of adaptive cross-modal blocks. Figure \ref{fig:adaptive} illustrates the architecture of the cross-modal attention and the adaptive cross-modal blocks. After obtaining the unimodal representations, we feed them into the adaptive cross-modal block to obtain the reinforcement features of both modalities. Here, we show the process of using audio modality to reinforce visual modality. The process of using visual modality to reinforce audio modality is the same. Specifically, we first make the audio modal pass through the Transformer encoder to perform intra-modal feature selection. The Transformer encoder is the same as the Cross-Transformer encoder, the difference lies in Equation 5:

\begin{equation}
    y^l = {\rm MSA}({\rm LN}(h_m^{l})) + h_m^{l} \quad m \in \{v,a\}.
\end{equation}

Here, the MSA operation calculates the dot product attention, where the queries, keys and values are all linear projections of the same tensor $h_m$, ${\rm MSA}(h_m) = {\rm Attention}(W^Q h_m, W^K h_m, W^V h_m)$. This operation enables the higher-order features of the audio modality to perform feature selection, making it more focused on features that have a greater impact on the outcome.

Then, we make the automatically selected features and the video modality perform efficient inter-modal interactions. The module accepts input for two modalities, which is called $\hat{X}_A \in \mathbb{R}^{d_f}$ and $\hat{X}_V \in \mathbb{R}^{C \times S \times H \times W}$. We obtain the mapping representations of the features for the two modalities by a linear projection. And then we process the two representations by $add$ and $tanh$ activation function. Finally, the fused representation $\hat{X}_o \in \mathbb{R}^{C \times S \times H \times W}$ is obtained through $softmax$. The specific formula is as follows:

\begin{equation}
    \hat{X}_q = tanh((W_v\hat{X}_V + b_v) + W_a\hat{X}_A) \in \mathbb{R}^{k \times S \times H \times W}
\end{equation}
\begin{equation}
    \hat{X}_o = (softmax(\hat{X}_q) \otimes \hat{X}_V) \oplus \hat{X}_V \in \mathbb{R}^{C \times S \times H \times W}
\end{equation}
in which $W_v \in \mathbb{R}^{k \times C}$ and $W_a \in \mathbb{R}^{k \times d_f}$ are linear transformation weights, and $b_v \in \mathbb{R}^{k}$ is the bias, where $k$ is a pre-defined hyper-parameter, and $\oplus$ represents the broadcast addition operation of a tensor and a vector. We can see that this operation is mainly to fuse the audio modality and the visual modality after feature selection by splicing to obtain the weight vector, and multiply the weight vector with the visual modality to achieve feature reinforcement. In this process, to ensure that the information of the  visual modality is not lost, we ensure the integrity of the original structural features of the visual modality through the residual structure. 

We use $\hat{X}_o^{a->v}$ to denote the features that use audio data to enhance the visual modality and $\hat{X}_o^{v->a}$ to denote the features that use visual data to enhance the audio modality. Finally, we splice the reinforced modalities to obtain the fused data $I = [\hat{X}_o^{v->a}, \hat{X}_o^{a->v}]$.

\subsection{Classification}

Finally, we use the fused data for emotion category prediction. The cross-entropy loss is used to optimize the model. The specific equation is shown as follows:

\begin{equation}
    {\rm prediction} = W_1I + b_1 \in \mathbb{R}^{d_{out}}
\end{equation}
\begin{equation}
    L = -\sum_{i} y_ilog(\hat{y}_i)
\end{equation}
where $d_{out}$ is the output dimensions of emotional categories, $ W_1 \in {R}^{d_{out}}$ is the weight vectors, $b_1$ is the bias, $y = \{y_1, y_2, ..., y_n\}^T$ is the one-hot vector of the emotion label, $\hat{y} = \{\hat{y}_1, \hat{y}_2, ..., \hat{y}_n\}^T$ is the predicted probability distribution, $n$ is the number of emotion categories.

\section{Experiments}

\subsection{Datasets}

In this paper, we use two mainstream datasets: RAVDESS and IEMOCAP. For simplicity, this study first focuses on the fusion of two modalities, visual and audio. Further, we extend the designed fusion block to cross-modal fusion of three modalities, visual, audio and text. Specifically, for the RAVDESS dataset, we use two modalities: visual and audio. For the IEMOCAP dataset, we use three modalities: visual, audio and text. The code will be publicly available after the paper is accepted.

\subsubsection{RAVDESS}

The Ryerson Audio-Visual Database of Emotional Speech and Song (RAVDESS) \cite{livingstone2018ryerson} is a multimodal emotion recognition dataset containing 24 actors (12 male, 12 female) of 1440 video clips of short speeches. The dataset is performed when the actors are told the emotion to be expressed, with high quality in terms of both video and audio recordings. Eight emotions are included in the dataset: neutral, calm, happy, sad, angry, fearful, disgust and surprised. We perform 5-fold cross-validation on the RAVDESS dataset to provide more robust results. We divide the 24 actors into a training and a test set in a 5:1 ratio. Since the actors' gender is represented by an even or odd number of actor IDs, we enable gender to be evenly distributed by rotating 4 consecutive actor IDs as the test set for each fold of cross-validation. The final accuracy reported is the average accuracy over the 5 folds.

\subsubsection{IEMOCAP}

IEMOCAP \cite{iemocap} is a multimodal emotion recognition dataset that contains 151 videos along with corresponding transcripts and audios. In each video, two professional actors conduct dyadic conversations in English. Its intended data segmentation consists of 2,717 training samples, 798 validation samples and 938 test samples. The audio and visual features are extracted at the sampling frequencies of 12.5 Hz and 15 Hz, respectively. Although the human annotation has nine emotion categories, following the prior works \cite{dai2020modalitytransferable}, we take four categories: neutral, happy, sad, and angry. Moreover, this is a multi-label task (e.g., a person can feel sad and angry at the same time).  We report the binary classification accuracy and F1 scores for each emotion category according to \cite{lv2021progressive}.

\subsection{Implementation details}

\subsubsection{Feature extraction of the RAVDESS dataset}

For the visual modality, we extract 30 consecutive images from each video. We crop the face region using the 2D face markers provided for each image and then resize to (224,224). Data augmentation is performed using random cropping, level flipping and normalization methods. For the audio modality, since the first 0.5 seconds usually do not contain sound, we trim the first 0.5 seconds and keep it consistent for the next 2.45 seconds. Following the suggestion of \cite{jin2015speech}, we extract the first 13 MFCC features for each cropped audio clip.

\subsubsection{Feature extraction of the IEMOCAP dataset}

For feature extraction of the text modality, we convert video transcripts into pre-trained Glove \cite{glove} model to obtain 300-dimensional word embeddings. For feature extraction of visual modality, we use Facet \cite{facenet} to represent 35 facial action units, which record facial muscle movements for representing basic and high-level emotions in each frame. For the audio modality, we use COVAREP \cite{covarep} for extracting acoustic signals to obtain 74-dimensional vectors.

The model is trained using the Adam optimizer \cite{kingma2014adam} with a learning rate of 0.001, and the entire training of the model is done on a single NVIDIA RTX 8000.

\subsection{Baselines}

For the audio-visual emotion recognition task, we implement multiple recent multimodal fusion algorithms as our baselines. We categorize them into the following:
\begin{itemize}
    \item Simple feature concatenation followed by fully connected layers based on \cite{ortega2019multimodal} and MCBP \cite{fukui2016multimodal} are two typical early fusion methods.
    \item Averaging and multiplication are the two standard late fusion methods that are adopted as the baselines.
    \item Multiplicative layer \cite{liu2018learn} is a late fusion method that adds a down-weighting factor to CE loss to suppress weaker modalities.
    \item MMTM \cite{joze2020mmtm} module allows slow fusion of information between modalities by adding to different feature layers, which allows the fusion of features in convolutional layers of different spatial dimensions.
    \item MSAF \cite{su2020msaf} module splits each channel into equal blocks of features in the channel direction and creates a joint representation that is used to generate soft notes for each channel across the feature blocks.
    \item MCA \cite{lv2021progressive} stands for cross-modal attention, and the module is a modification of self-attention, with Q as one modality and K and V as another modality to obtain the reinforcement of the modality, which is the current mainstream fusion method.
\end{itemize}

For multimodal emotion recognition tasks, we compare the proposed approach with the existing state-of-the-art methods:

\begin{itemize}
    \item Early Fusion LSTM (EF-LSTM) and Late Fusion LSTM (LF-LSTM) simply concatenate features at input and output level, which apply LSTM \cite{6795963} to extract features and infer prediction.
    \item Recurrent Attended Variation Embedding Network (RAVEN) \cite{2019work} and Multimodal Cyclic Translation Network (MCTN) \cite{mctn} are joint representation fusion methods based on temporal modeling.
    \item Multimodal Transformer (MulT) \cite{tsai2019multimodal}, Low Rank Fusion based Transformers (LMF-MulT) \cite{sahay2020low} and Progressive Modality Reinforcement (PMR) \cite{lv2021progressive} are all inter-modal complementary fusion methods based on cross-modal attention, with PMR being the state-of-the-art model.
\end{itemize}

\subsection{Comparison to state-of-the-art methods}

Table \ref{tab:avg} shows the accuracy comparison of the proposed method with baselines on RAVDESS dataset. From the table \ref{tab:avg}, we can see that our model achieves an accuracy of 76.76\% reaching the state-of-the-art. \textbf{1)} The unimodal performance of TACFN is 62.99\% and 56.63\% for visual and audio, respectively, and the accuracy is 76.76\% after the adaptive cross-modal blocks, which is an improvement of more than 13.77\%. As can be seen, the adaptive cross-modal block learns the complementary information of both. It learns information that is present in the audio but not in the visual, thus giving the visual representation more semantic information of the audio modality. \textbf{2)} Compared with early fusion methods, our method has more than 5\% accuracy improvement, which shows that finding the association between  visual and audio modalities is a difficult task in the early stage. Compared to the late fusion methods, we find that the unimodal feature splicing obtained in the late stage will get 73.50\% performance. What's more, our proposed TACFN has 3.64\% improvement over MMTM and 1.9\% improvement over the best-performing MSFA. \textbf{3)} We obtain the reinforcement features fusing the other modality using two cross-modal attentions separately. The experimental results show an accuracy of 74.58\%, while our accuracy is 76.76\%. TACFN is more effective under the same experimental setup. The reason that cross-modal attention does not capture complementary features well is considered to be due to the fact that audio modality is not entirely helpful for visual modality, while features that often reinforce visual modality may only contain a portion of it. Based on this, we have designed the cross-modal block to focus more effectively and adaptively on information that is more useful for the current modality, i.e., effective complementary information.

We also apply the model to the IEMOCAP dataset. Table \ref{tab: iemocap} shows the results. We use adaptive cross-modal blocks to achieve complementary learning, i.e., we use audio and text fusion modalities to obtain weight information to reinforce text modalities, and visual and text fusion modalities to obtain weight information to reinforce text modalities. \textbf{1)} MulT utilize cross-modal attention to achieve complementary learning. Our model outperforms MulT in all metrics. \textbf{2)} We also compare with the state-of-the-art PMR, and the results achieved comparable levels. Surprisingly, TACFN has the lowest number of parameters.

\begin{tabhere}
    \centering
    \setlength{\tabcolsep}{0.8mm}
    \caption{ Comparison between multimodal fusion baselines and ours for emotion recognition on RAVDESS.}
    \label{tab:avg}
    \begin{tabular}{cccc}
        \toprule
        Model & Fusion & Accuracy(\%) & Improve(\%) \\ 
        \midrule
        3D RexNeXt50 (Vis.) & - & 62.99 & - \\
        1D CNN (Aud.) & - & 56.53 & - \\
        \hline
        Averaging & Late & 68.82 & + 5.83 \\
        Multiplicative $\beta$=0.3 & Late & 70.35 & + 7.36 \\
        Multiplication & Late & 70.56 & + 7.57 \\
        Concat + FC & Late & 73.50 & + 10.51 \\
        Concat + FC \cite{ortega2019multimodal} & Early & 71.04 & + 8.05 \\
        MCBP \cite{fukui2016multimodal} & Early & 71.32 & + 8.33 \\
        MMTM \cite{joze2020mmtm} & Model & 73.12 & + 10.13 \\
        MSAF \cite{su2020msaf} & Model & 74.86 & + 11.87 \\
        \hline
        MCA \cite{lv2021progressive} & Model & 74.58 & + 11.59 \\
        \textbf{TACFN  (Ours)} & Model & \textbf{76.76} & \textbf{+ 13.77} \\
        \bottomrule
    \end{tabular}%
\end{tabhere}

\subsection{Ablation study}

\subsubsection{Effectiveness of adaptive cross-modal blocks}

Table \ref{tab:pool} shows the ablation experiments on the RAVDESS dataset. To verify the effectiveness of the adaptive cross-modal blocks, we obtain the final sentiment by simply splicing the high-level semantic features of the two modalities. The experimental results show that our cross-modal block leads to a performance improvement of more than 3\% with only a 0.4M increase in the number of parameters, which indicates that efficient complementary information from both modalities can have a large impact on the final decision.

We further explore the validity of the internal structure of the adaptive cross-modal blocks. In the adaptive cross-modal blocks, the self-attention mechanism and the residual structure play an important role in the performance of the model. We detach the self-attention mechanism and the residual structure separately which can be seen that self-attention brings more than 3\% impact on the final result. This indicates that the audio semantic features we obtained contain redundant information and can be selected by the self-attention mechanism for feature selection to make it efficient and adaptive for inter-modal interaction. In addition, we also see that the residual structure has less impact on the final results, suggesting that the inclusion of the residual structure helps to ensure that the loss of visual features is minimized during the interaction.

\begin{table*}[!t]
    \centering
    \setlength{\tabcolsep}{1.5mm}
    \caption{Comparison on the IEMOCAP dataset under both word-aligned setting and unaligned setting. The performance is evaluated by the binary classification accuracy and the F1 score for each emotion class. TACFN achieves comparable and superior performance with only 0.34M parameters.}
    \begin{tabular}{ccccccccccc}
    \toprule
    \multirow{2}*{Method} & \multirow{2}*{Fusion} & \multirow{2}*{\#Param} & \multicolumn{2}{c}{Happy} & \multicolumn{2}{c}{Sad} & \multicolumn{2}{c}{Angry} & \multicolumn{2}{c}{Neutral} \\
    & & & $Acc(\%)$ & $F1(\%)$ & $Acc(\%)$ & $F1(\%)$ & $Acc(\%)$ & $F1(\%)$ & $Acc(\%)$ & $F1(\%)$ \\ 
    \hline
    EF-LSTM  & Early & 0.10M & 76.2 & 75.7 & 70.2 & 70.5 & 72.7 & 67.1 & 58.1 & 57.4  \\
    LF-LSTM & Late  & 0.17M &72.5 & 71.8 & 72.9 & 70.4 & 68.6 & 67.9 & 59.6 & 56.2  \\
    RAVEN \cite{2019work} & Model  & 1.20M &77.0 & 76.8 & 67.6 & 65.6 & 65.0 & 64.1 & 62.0 & 59.5  \\
    MCTN \cite{mctn} & Model  & 0.97M & 80.5 & 77.5 & 72.0 & 71.7 & 64.9 & 65.6 & 49.4 & 49.3  \\
    MulT \cite{tsai2019multimodal} & Model  & 1.07M & 84.8 & 81.9 & 77.7 & 74.1 & 73.9 & 70.2 & 62.5 & 59.7  \\
    LMF-MulT \cite{sahay2020low}  & Model  & 0.86M &85.6 & 79.0 & 79.4 & 70.3 & 75.8 & 65.4 & 59.2 & 44.0  \\
    PMR \cite{lv2021progressive} & Model &  2.15M & \textbf{86.4} & \textbf{83.3} & 78.5 & 75.3 & 75.0 & 71.3 & \textbf{63.7} & \textbf{60.9}  \\
    \hline
    \textbf{TACFN (Ours)} & Model & \textbf{0.34M} & 85.7& 82.5 & \textbf{79.4} & \textbf{75.6} & \textbf{76.0} & \textbf{71.7} & 63.6 & 60.5    \\
    \bottomrule
    \end{tabular}
    \label{tab: iemocap}
\end{table*}

More, we integrate the audio modality into the visual modality as the final fusion result in our model design, denoted as $V$-\textgreater$A$. We have compared the results of integrating visual modality into audio modality, denoted as $A$-\textgreater$V$. We find that there is a 1.6\% difference between them. We splice them together to bring a 1\% improvement.

\subsubsection{The validity of each category}

We report the accuracy of unimodal and TACFN for each class on the RAVDESS dataset separately. The results are shown in Table \ref{tab: ravdess}. \textbf{1)} For the visual modality, the expressions ``Happy", ``Angry" and ``Surprised" are easier to distinguish than the audio modality. \textbf{2)} For the audio modality, ``Sad" and ``Fearful" have higher performance. \textbf{3)} It can be seen that after the adaptive cross-modal blocks, the accuracy of each class has improved compared to the unimodal ones. ``Neutral" has the lowest accuracy while ``Sad" and ``Fearful" have the most significant improvement of about 10\%. We believe that the visual modality has gained complementary information from the audio modality, resulting in a higher performance.

\begin{tabhere}
    \setlength{\tabcolsep}{1.3mm}
    \fontsize{9}{11}\selectfont
    \centering
    \caption{\fontsize{9}{11}\selectfont  Ablation study on the RAVDESS dataset.
    }
    \label{tab:pool}
    \begin{tabular}{lcccc}
        \toprule
        Model & Accuracy & \#Params \\ 
        \midrule
        TACFN & 76.76 & 26.30M \\
        \hline
        w/o Adaptive Cross-modal Blocks & 73.50 & 25.92M \\
        \hline
        w/o self-attention & 73.86 & 26.05M \\
        w/o residual & 76.33 & 26.30M \\
        \hline
        V-\textgreater A Adaptive Cross-modal & 75.15 & 25.67M \\
        A-\textgreater V Adaptive Cross-modal & 75.76 & 26.30M \\
        \bottomrule
    \end{tabular}%
\end{tabhere}

\begin{tabhere}
    \centering
    \setlength{\tabcolsep}{5mm}
    \caption{ The accuracy of each class in the RAVDESS dataset.}
    \begin{tabular}{cccc}
    \toprule
    Label & Audio & Visual & TACFN \\
    \hline
    Neutral  & 54.2  & 58.6  & 65.6   \\
    Calm  & 63.4 & 66.5 & 71.2  \\
    Happy  & 57.3  & 68.1  & 70.2   \\
    Sad & 72.0  & 64.9  & 81.4   \\
    Angry  & 68.3  & 77.2  & 87.0   \\
    Fearful & 73.4  & 68.8  & 86.3  \\
    Disgust  & 61.5  & 69.5 & 73.2   \\
    Surprised   & 74.1  & 76.3  & 81.7   \\
    \bottomrule
    \end{tabular}
    \label{tab: ravdess}
    \vspace{-0.2in}
\end{tabhere}

\section{Conclusion}

In this paper, we propose an innovative transformer-based adaptive multimodal fusion network. We divide this network into two steps: unimodal representation and multimodal fusion. The two core issues of reducing redundant features and enhancing complementary features are mainly considered in multimodal fusion. For reducing redundant features, we use a self-attention mechanism to enable one modality to perform intra-modal feature selection, and the selected features can be adaptively interacted with another modality in an efficient inter-modal manner. For boosting complementary features, we fuse the selected modality with another modality by splicing to obtain a weight vector, and multiply the weight vector with another modality to achieve feature reinforcement. We apply the model to RAVDESS and IEMOCAP datasets, and the experimental results show that our proposed fusion method is more effective. Compared with other models, our approach delivers a significant performance improvement in emotion recognition from the fusion strategy based on the same unimodal representation learning.

	\section*{Dates}
	\noindent Received: 7 July 2023; Accepted:X XXX 2023;
	
	\noindent Published online: X September 2023

	\fontsize{8pt}{10.2pt}\selectfont

	\bibliographystyle{unsrt}
	
	\bibliography{Reference}
	
	\end{multicols}
\newpage

\end{document}